\documentclass[runningheads]{llncs}
\usepackage{tablefootnote}
\usepackage{graphicx}
\usepackage{todonotes}
\usepackage{booktabs}
\usepackage{multirow}
\usepackage{hyperref}
\usepackage{subcaption}
\usepackage[export]{adjustbox}
\usepackage{xcolor}
\definecolor{blue}{HTML}{4472C4}
\definecolor{darkblue}{HTML}{51607A}
\definecolor{green}{HTML}{70AD47}
\definecolor{orange}{HTML}{ED7D31}
\definecolor{brown}{HTML}{5D3709}
\definecolor{purple}{HTML}{7030A0}
\definecolor{yellow}{HTML}{FFC000}
\definecolor{teal}{HTML}{00CC99}
\definecolor{pink}{HTML}{CC0099}
\definecolor{grey}{HTML}{979A9A}
\definecolor{darkred}{HTML}{a93226}

\DeclareUnicodeCharacter{24BC}{{%
  \fontencoding{TS1}%
  \selectfont s%
}}
\DeclareUnicodeCharacter{24C5}{{%
  \fontencoding{TS1}%
  \selectfont s%
}}

\begin{document}

\title{Key-value information extraction from full handwritten pages}

%
%

\author{Solène Tarride\inst{1}\orcidID{0000-0001-6174-9865} \and
Mélodie Boillet\inst{1,2}\orcidID{0000-0002-0618-7852} \and
Christopher Kermorvant\inst{1,2}\orcidID{0000-0002-7508-4080} 
}

\authorrunning{S. Tarride et al.}
%
\institute{TEKLIA, Paris, France \and
LITIS, Normandy University, Rouen, France}
\maketitle              
%


\begin{abstract}
We propose a Transformer-based approach for information extraction from digitized handwritten documents.
Our approach combines, in a single model, the different steps that were so far performed by separate models: feature extraction, handwriting recognition and named entity recognition.
We compare this integrated approach with traditional two-stage methods that perform handwriting recognition before named entity recognition, and present results at different levels: line, paragraph, and page.
Our experiments show that attention-based models are especially interesting when applied on full pages, as they do not require any prior segmentation step.
Finally, we show that they are able to learn from key-value annotations: a list of important words with their corresponding named entities.
We compare our models to state-of-the-art methods on three public databases (IAM, ESPOSALLES, and POPP) and outperform previous performances on all three datasets. 

\end{abstract}

\section{Introduction}

Although machine learning and deep learning techniques are nowadays commonly used in the field of automatic processing of historical documents\cite{lombardi2020}, scientific work still often focuses on some specific processing steps in isolation. It is common to develop models either for page analysis or line detection, for handwriting recognition or for information extraction. Processing chains are still often developed as a sequence of these steps independently. However, these processing chains suffer from several drawbacks. Firstly, errors accumulate along the chain: if the line detection step is bad, write recognition will be highly impacted and information extraction impossible. On the other hand, the implementation of these chains and their maintenance is complex: each step requires specific skills and annotated data for each model and any update of a part of the chain has an impact on all downstream processes. Finally, the different modules are developed independently and there is no global optimization of the processing chain. For all these reasons, the development of models allowing the extraction of information directly from the image, by an end-to-end approach, with a single model, would be very beneficial.    

As far as automatic recognition is concerned, three main types of projects are currently being carried out on collections of historical documents, depending on the intended use. The first type of project aims to carry out a complete transcription of the documents to allow full-text searches \cite{muehlberger2019,vidal2020}. The processing chain then focuses on the page analysis stage to extract a maximum number of lines of text and the handwriting recognition stage to best recognize the text. The result of the processing is then exploited thanks to a search engine that allows queries to be made and documents to be identified according to their content. The second type of processing aims to produce electronic editions of documents\cite{kiessling2019}. In this case, the emphasis is obviously on the quality of the recognition, but also the fidelity to the text of the document and the reading order. The result of the automatic processing is in this case always submitted to the correction of an expert before publication. The last type of project aims at extracting information from documents in order to populate a database with the information they contain \cite{tarride2023}. In addition to the document analysis and handwriting recognition stages, these projects also incorporate an information extraction stage, often in the form of named entity extraction. It is this third type of project, the most complex in its implementation, that we are interested in  this work.

Information extraction chains for historical handwritten documents are usually composed of the following steps: line detection or document layout analysis (DLA), handwriting recognition (HTR) and named entity extraction (NER). In this paper, we first reconsider the possibility of combining the HTR and NER models into a single model. Then we study whether it is possible to extend this model to the processing of a complete page without going through a line detection step. Finally, we show that it is possible to go even further and train a single model for the extraction of target information, of the key-value type, without going through an explicit transcription.

The rest of this paper is organized as follows. In Section 2, we review the state-of-the-art for information extraction in handwritten documents. We describe our methodology and experiments in Section 3. The experimental results are presented and analyzed in Section 4. Finally, in Section 5 we discuss the conclusions and outline future works.

\section{Related work}

Recent advances in computer vision and natural language processing have led to major breakthroughs in the field of automatic document understanding. 
Deep learning-based systems are now capable of automatically extracting relevant information from historical documents. 
Interest in this field has been encouraged by the emergence of competitions, such as the Information Extraction on Historical Handwritten Records competition \cite{iehhr2017} on the ESPOSALLES database \cite{Esposalles}, as well as the publication of named entity recognition annotations for other databases, such as IAM-NER \cite{Tuselmann2021-NER-seq} and POPP \cite{Constum-POPP}.

Two main approaches exist to address automatic information extraction from handwritten documents:
\begin{itemize}
    \item \textbf{Sequential approaches} consist in dividing the problem into two successive tasks: handwritten text recognition, and named entity recognition;
    \item \textbf{Integrated approaches} consist in combining text and named entity recognition in a single-step.
\end{itemize}

Each of these approaches can work at several levels: either on words, lines, paragraphs, or directly on full pages. Segmentation-based systems work on pre-segmented text zones (words, lines, or paragraphs), while segmentation-free systems work directly on full pages. Performing handwriting recognition on smaller zones is usually easier to achieve, but requires a prior segmentation step. As opposed, handwriting recognition on full pages is more challenging (memory management, reading order), but does not require any prior segmentation.

\subsection{Sequential approaches}

In sequential approaches, HTR is performed first, then, NER is applied on recognized text. Note that HTR and NER can be applied at different levels: HTR is usually performed at line-level, and NER at paragraph or page-level.

\subsubsection{Segmentation-based systems}

Five systems were introduced during the ICDAR2017 Competition on Information Extraction in Historical Handwritten Records \cite{iehhr2017} on ESPOSALLES. 
Most participants used CRNN trained with CTC to recognize handwritten text. Named entity recognition was then performed using logical rules based on regular expressions or CRF tagging. Other methods were proposed after the competition. 

Prasad et al. \cite{Prasad2018-IEHHR-benchmark} propose a two-stage system combining a CRNN-CTC neural network for HTR on text line images, followed by a BLSTM layer over the feature layer for NER. 

Tuselmann et al. \cite{Tuselmann2021-NER-seq} also introduce a two-stage system for information extraction that combines a Transformer model \cite{Kang2018-HTR} for HTR on word images, and a LSTM-CRF model with word embeddings obtained using a pre-trained RoBERTa for NER. They highlight the advantages of two-stage methods for information extraction, as these methods yield state-of-the art results and are easy to improve using post-processing techniques, close dictionary, or pre-trained embeddings. 

Monroc et al. \cite{Monroc2022-NER-opensource} compare different off-the-shelf NER libraries on handwritten historical documents: SpaCy \cite{spacy}, FLAIR \cite{flair}, and Stanza \cite{stanza}. They perform experiments on three datasets in an end-to-end setting, and study the impact of text line detection and text line recognition on NER performances. Their results highlight that line detection errors have a greater impact than handwriting recognition errors. This conclusion suggests that working directly on pages could prevent segmentation errors from impacting the final entity recognition.

\subsubsection{Segmentation-free systems}

To the best of our knowledge, no system performing sequential HTR and NER at page-level has been proposed so far. However, many segmentation-free HTR models working directly at page-level \cite{OrigamiNet,VAN,DAN} have been introduced recently. Any of these models could easily be combined with off-the-shelf NER libraries for segmentation-free information extraction.

\subsection{Integrated approaches}

Integrated approaches combine HTR and NER in a single step by modeling named entities with special tokens. This can be achieved with or without prior segmentation.

\subsubsection{Segmentation-based systems}

Toledo et al. \cite{Toledo2019-context} and Rowtula et al. \cite{Rowtula2018-POS} introduce models that work at word-level. Their systems are able to recognize and classify word images into semantic categories. 

Both Carbonell et al. \cite{Carbonell2018-tags} and Tarride et al. \cite{Tarride2022-IEHHR-Attention} propose neural networks that predict characters and semantic tags from line images, respectively, using a CRNN model trained with CTC and an attention-based network. Both of these studies suggest that working on records would allow the model to capture more contextual information.

In \cite{Constum-POPP}, the authors use the same approach on French census images from the POPP dataset, as they predict text characters and special tokens for empty cells and column separators. Although, this dataset does not directly include named entities, each word is linked to a specific column and can be seen as a named entity (name, surname, date of birth, place of birth...). 

Finally, Rouhou et al. \cite{Rouhou2021-IEHHR-Transformer} are the first to introduce a Transformer model for combined HTR and NER at record-level on the ESPOSALLES database. They highlight the interest of performing this task on records to benefit from more contextual information. As each page contains several records, this model still requires record segmentation. Moreover, they use a special token for line breaks, as they observe this improves performance. 

\subsubsection{Segmentation-free systems}

Carbonell et al. \cite{Carbonell2020-dla+htr+ner-words} are the first to propose a model that works directly at page-level on ESPOSALLES. Their system is able to jointly learn word bounding boxes, word transcription and word semantic category on ESPOSALLES. However, a major limitation of this method is that it requires word bounding boxes during training. 

The Transformer proposed by Rouhou et al. \cite{Rouhou2021-IEHHR-Transformer} could be applied to full pages in its current stage, although this task has not been tackled by the authors. 

Finally, the Document Attention Network (DAN) \cite{DAN} is able to recognize text on full pages with reading order. It is based on the Transformer architecture and jointly learns characters and special tokens that represent layout information. It is likely that this method is also able to recognize named entities, or in other words, tokens that are not spatially localized but have a semantic meaning. However, the authors did not perform any experiments on named entity recognition. 



\subsection{Discussion}

The literature review opens up three main questions that are discussed in the following.

\subsubsection{What is the best approach for information extraction?}
Although this question has been well studied in the past, no consensus has been reached. On the one hand, researchers have shown the interest of sequential methods which can be optimized at every stage (with a language model, a dictionary, or pre-trained embeddings) \cite{Tuselmann2021-NER-seq,Monroc2022-NER-opensource}. On the other hand, the advantages of integrated methods have also been demonstrated \cite{Tarride2022-IEHHR-Attention}, notably because they benefit from shared contextual features and avoid cascade errors.

\subsubsection{Can we extract relevant information from full pages?}
Different methods were designed to work at different levels, some of them requiring prior segmentation of text lines or paragraphs. However, in real-world scenarios, text areas are not known and must therefore be detected automatically, which can introduce segmentation errors. It has been established that segmentation errors have a greater impact on information extraction than handwriting recognition errors \cite{Monroc2022-NER-opensource}.
Recently, Transformers have proved their ability to learn from paragraphs and pages \cite{DAN,Rouhou2021-IEHHR-Transformer}, enabling segmentation-free information extraction. Learning directly from pages increases the task difficulty, but avoids the need for prior segmentation. Moreover, working directly on pages makes it possible to benefit from a larger context \cite{Rouhou2021-IEHHR-Transformer}.

\subsubsection{Are integrated models able to learn from key-value annotations?}
As sequential approaches rely on HTR, they require the entire transcription before retrieving named entities. However, integrated methods could potentially learn from key-value annotations, which corresponds to a list of words with their corresponding named entities. In this scenario, ground-truth is also easier and faster to produce, as annotators would only have to annotate important words as well as their semantic category. This approach could also be applied in a lot of practical applications where full transcriptions are not available, such as genealogical crowdsourced information (civil status, or personal records). This question has not been studied yet in the context of information extraction.

In the next section, we describe the experiments designed to address these three questions. 

\section{Methodology and experiments}

In this section, we introduce the datasets used during our experiments, present our methodology, and describe the different experiments conducted in this study.

\subsection{Datasets}

During our experiments, we worked on three public datasets of different kind.

\begin{figure}
    \centering
    \begin{subfigure}[b]{0.32\textwidth}
         \includegraphics[height=0.3\textheight, center]{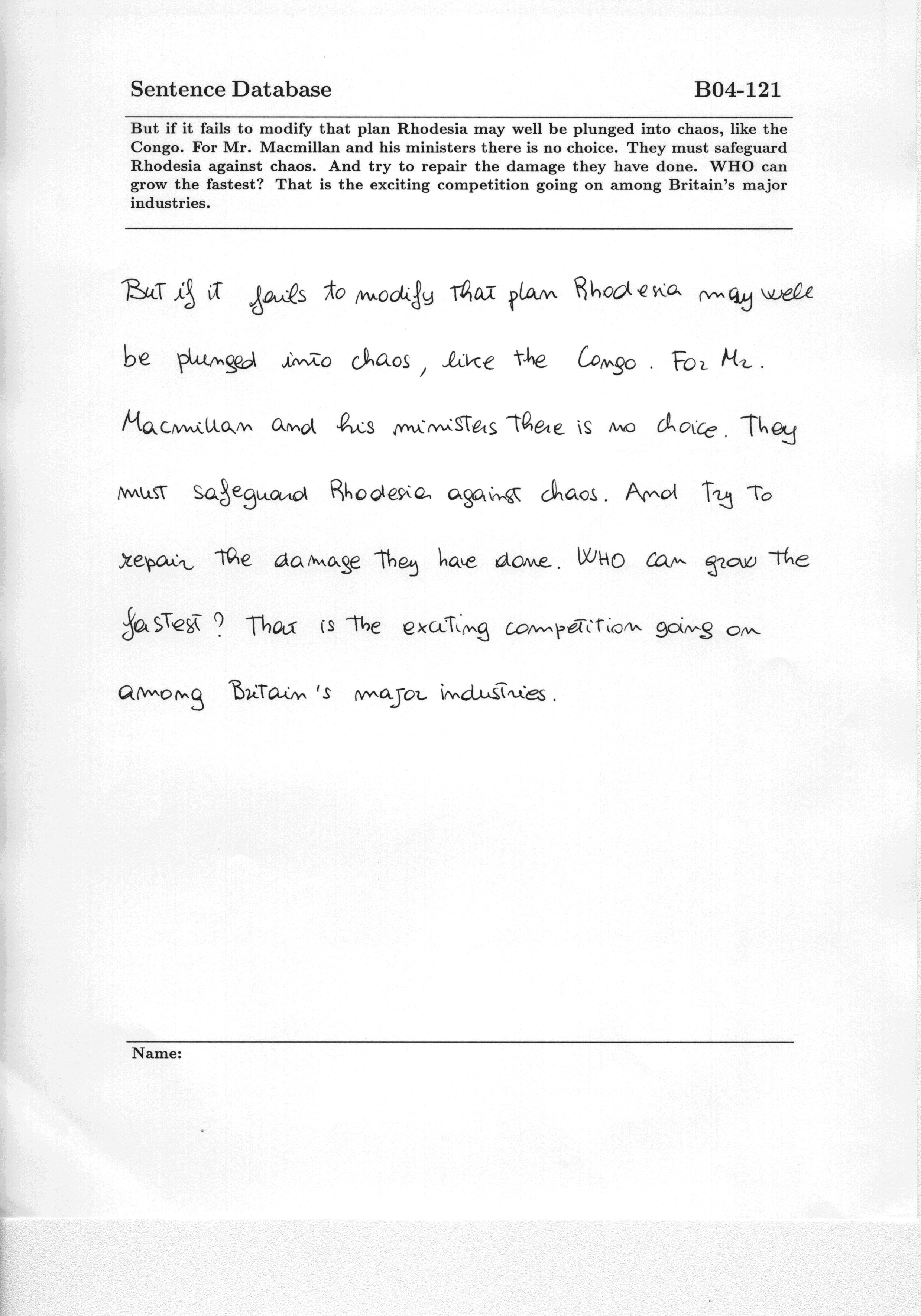}
         \caption{IAM \cite{IAM}}
         \label{fig:iam}
     \end{subfigure}     
    \begin{subfigure}[b]{0.32\textwidth}
         \includegraphics[height=0.3\textheight, center]{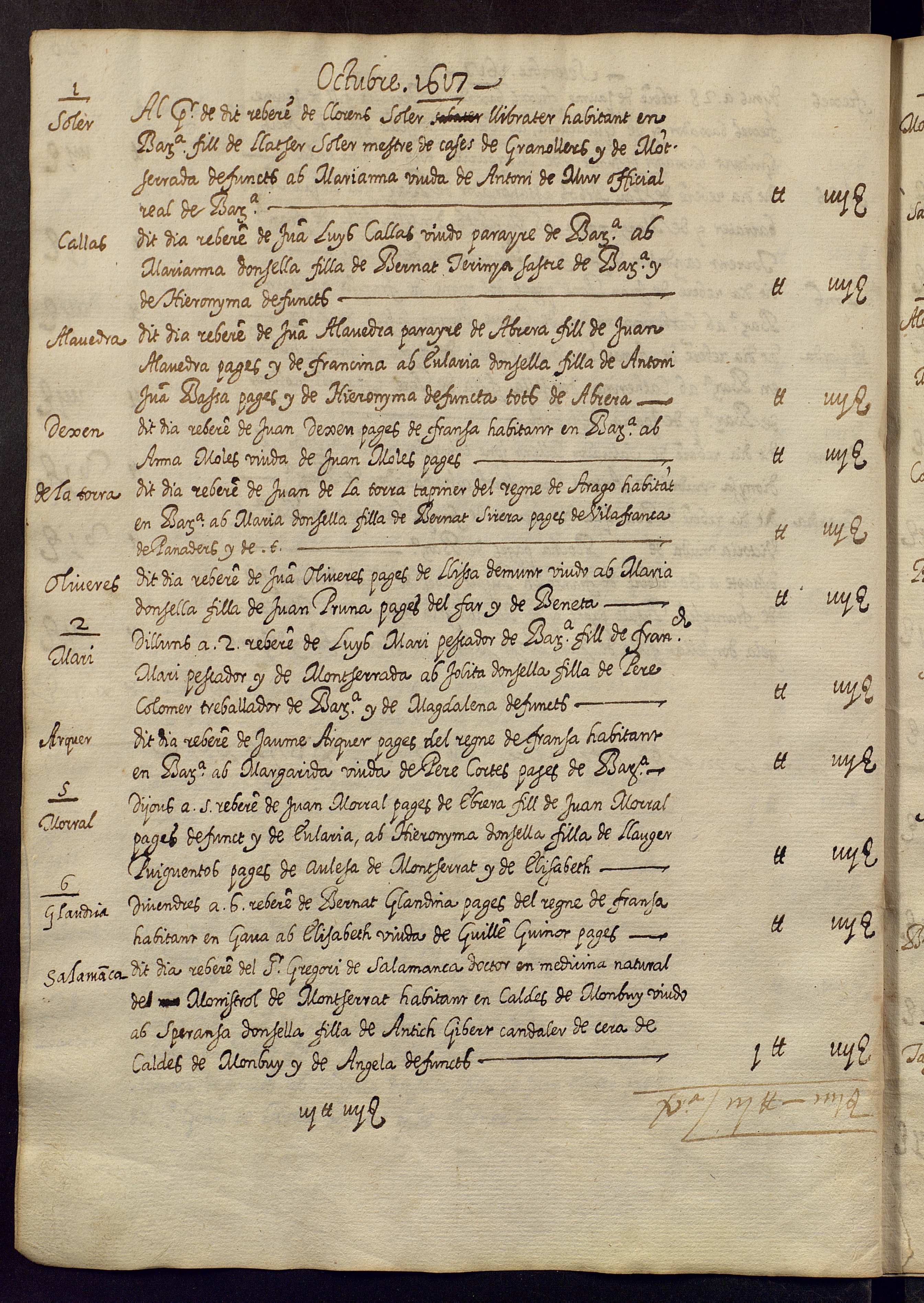}
         \caption{ESPOSALLES \cite{Esposalles}}
         \label{fig:esposalles}
     \end{subfigure}   
    \begin{subfigure}[b]{0.32\textwidth}
         \includegraphics[height=0.3\textheight, center]{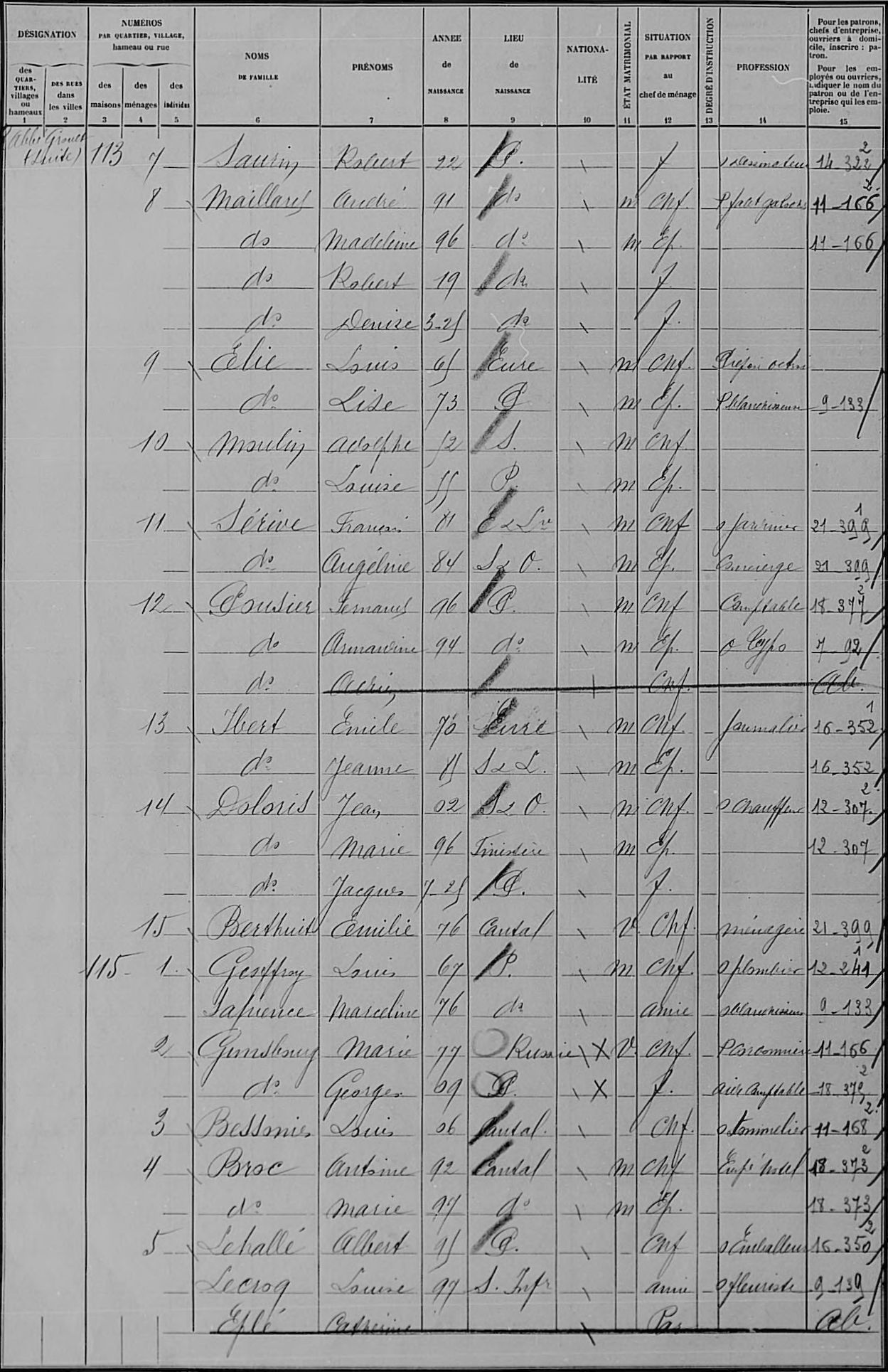}
         \caption{POPP \cite{Constum-POPP}}
         \label{fig:popp}
     \end{subfigure}   
    \caption{Examples of pages from the three datasets used in this work}
    \label{fig:datasets}
\end{figure}

\subsubsection{IAM}
The IAM dataset \cite{IAM} is composed of modern documents written in English by 500 writers. It includes 747 training pages with corresponding transcriptions. NER annotations have been made available by Tüselmann et al. \cite{Tuselmann2021-NER-seq}. A page from IAM is presented in Figure \ref{fig:iam}. 

For our experiments, we use the RWTH split with 18 entities: Cardinal, Date, Event, FAC, GPE, Language, Law, Location, Money, NORP, Ordinal, Organization, Person, Percent, Product, Quantity, Time and Work of art. The details of this split are provided in the appendix. 
Less than 10\% of words are associated to an entity. Due to the large number of classes, some entities have very few examples in the training set.
We perform experiments at two levels: text line and page. When working on pages, we remove the header so that the model does not see the printed transcription instruction.

\subsubsection{ESPOSALLES}

The ESPOSALLES dataset \cite{iehhr2017} is a collection of historical marriage records from the archives of the Cathedral of Barcelona. The corpus is composed of 125 pages. Each document is written in old Catalan by a single writer. It includes 125 pages with word, line and record segmentations. The details of this split are provided in the appendix. A page from ESPOSALLES is presented in Figure \ref{fig:esposalles}. 

Each word is transcribed and labeled with a semantic category (name, surname, occupation, location, state, other) and a person (husband, wife, husband's father, husband's mother, wife's father, wife's mother, other person, none). More than 50\% of words are associated to an entity. As there is no validation set, we keep 25\% of training pages for validation.
We perform experiments at three levels: text line, record, and page. 

\subsubsection{POPP}

The POPP dataset contains tabular documents from the 1926 Paris census whose statistics are detailed in Table \ref{tab:popp_stats}. It contains 160 pages written in French, each page contains 30 lines. A page from POPP is presented in Figure \ref{fig:popp}.

Each row is divided in 10 columns: surname, name, birthdate, birthplace, nationality, civil status, link, education level, occupation, employer. In our experiments, we use the column name as a named entity. As a consequence, 100\% words are associated to an entity.
We perform experiments at two levels: text line and page.

\begin{table}[]
    \centering
    \caption{Statistics of the POPP dataset}
    \label{tab:popp_stats}
    \begin{subtable}{0.42\linewidth}
        \centering
        \caption{Pages, lines, words, and entities by split}
             \begin{tabular}{lrrr}
            \toprule
             & Train & Validation & Test \\
             \midrule
             Pages & 128 & 16 & 16 \\
             Lines & 3,837 & 480 & 479 \\
             Words & 29,581 & 3,681 & 3,569 \\
             Entities & 29,581 & 3,681 & 3,569 \\
             \bottomrule
        \end{tabular}
    \end{subtable}
    \hfill
    \begin{subtable}{0.49\linewidth}
        \centering
        \caption{Entities by split}
            \begin{tabular}{lrrr}
            \toprule
            & Train & Validation & Test \\
            \midrule
            Surname & 3,100 & 392 & 375 \\
            First name & 3,853 & 476 & 478 \\
            Birthdate & 3,824 & 469 & 466 \\
            Location & 4,789 & 600 & 584 \\
            Nationality & 283 & 17 & 30 \\
            Civil status & 2,277 & 292 & 225 \\
            Link & 3,667 & 449 & 412 \\
            Education level & 25 & 4 & 12 \\
            Occupation & 4,488 & 529 & 535 \\
            Employer & 3,275 & 453 & 452 \\
           \bottomrule
        \end{tabular}
    \end{subtable}
\end{table}

\subsection{Methods}

Three methods are introduced and compare in this work.


\subsubsection{Two-stage workflow}
The first method is a traditional two-stage workflow for information extraction that combines two steps. First, an HTR system is applied for text recognition on line-level images, then, SpaCy\footnote{\url{https://spacy.io}} \cite{spacy} is used for named entity recognition. We compare two systems for the HTR task: PyLaia \cite{PyLaia} and DAN \cite{DAN}.

\begin{itemize}
    \item PyLaia\footnote{\url{https://github.com/jpuigcerver/PyLaia}} is an open source model for handwritten text recognition. It combines 4 convolutional layers and 3 recurrent layers, and is trained with the CTC loss function. The last layer is a linear layer with a softmax activation function that computes probabilities associated with each character of the vocabulary. We use early stopping to avoid overfitting: the training is stopped after 50 epochs without improvement. PyLaia is trained on text line images.\\
    \item DAN\footnote{\url{https://github.com/FactoDeepLearning/DAN}} is an open source attention-based Transformer model for handwritten text recognition that can work directly on images of paragraph or page. It is trained with the cross-entropy loss function. The last layer is a linear layer with a softmax activation function that computes probabilities associated with each character of the vocabulary. For each dataset, we train DAN on zones with the strongest semantic consistency: on records for ESPOSALLES, on pages for IAM, and lines for POPP. \\
\end{itemize}

For NER, we use SpaCy, a production-oriented NLP library that includes transformer-based pipelines with support of English (for IAM), Catalan (for ESPOSALLES), and French (for POPP). Like DAN, SpaCy is trained on records for ESPOSALLES, pages for IAM, and lines for POPP. For ESPOSALLES, we train two SpaCy models: one for the \textit{category} label and one for the \textit{person} label.
Comparing two HTR systems with the same SpaCy model allows us to study the impact of handwriting recognition errors on the overall performance. 

\subsubsection{Integrated workflow}
The second method consists in training a model to recognize directly characters and NER tokens. \\
We train DAN models for this task, later referred to as \textit{HTR+NER}. The model is trained at different levels to evaluate the impact of context: on lines and pages for IAM, on lines, records and pages for ESPOSALLES, on lines and pages for POPP. 
NER tokens are considered like characters by the network and are localized before relevant words, as illustrated in Table \ref{tab:partial_tr}. 
For ESPOSALLES, we use a unique tag combining the \textit{category} and \textit{person} information (ex: \texttt{<name\_wife>Maria}), as we found out that using two separate tags led to poorer performance. This observation is consistent with the findings of Carbonell et al. \cite{Carbonell2018-tags} and Rouhou et al. \cite{Rouhou2021-IEHHR-Transformer}. 
Finally, we also trained DAN with curriculum learning, e.g. trained for \textit{HTR} and fine-tuned for \textit{HTR+NER} and found out that the network reach similar performance. For clarity, we only provide results without curriculum learning.

\subsubsection{Integrated workflow with key-value annotations}

Our last experiment consists in training DAN on key-value annotations, so as to only predict relevant information with the relevant text and the corresponding named-entity. This task is referred to as \textit{Key-value HTR+NER} in the rest of the article.
To achieve this, words that are not linked to any entities are removed from transcriptions, as illustrated in Table \ref{tab:partial_tr}. As a result, the model must learn to directly extract important words with their named entities, and ignore any other word. 
In this scenario, the two-stage approach cannot be used, as the full transcription is not available. 
This task is very challenging on IAM, as 90\% of words are not linked to any entities, and more than 5\% of pages do not have any entities. As a result, the training data is very sparse. 
The task is easier for ESPOSALLES, as 50\% of words are linked to an entity. 
Finally, in POPP, every word is related to a named entity, so the \textit{HTR+NER} and \textit{Key-value HTR+NER} tasks are the same.

\begin{table}[!ht]
    \centering
    \caption{Example of different transcriptions of the same record from the Esposalles database. Each transcription is used for a different task. \textit{HTR}: the model predicts characters, \textit{HTR+NER}: the model predicts characters and NER tokens, \textit{Key-value HTR+NER}: the model predicts characters and NER tokens only for relevant words, ignoring words that are not associated with NER tokens.}
    \label{tab:partial_tr}
    \begin{tabular}{lp{9cm}}
    \multicolumn{2}{c}{\raisebox{-\totalheight}{\includegraphics[width=\textwidth]{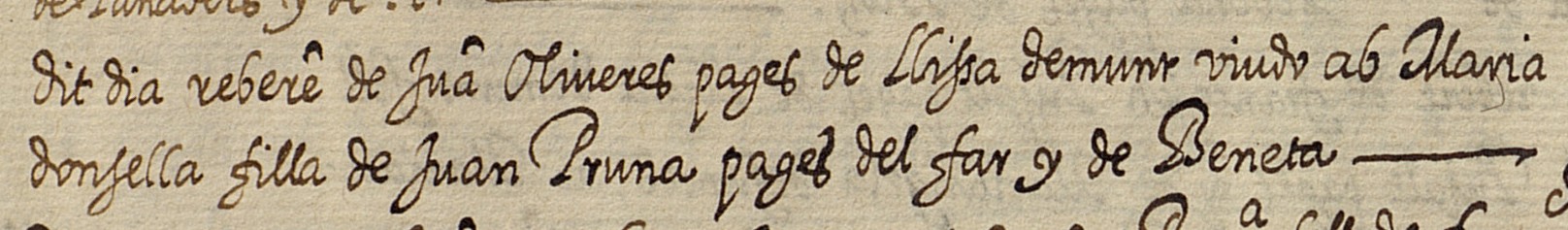}}} \\
    \\
    \toprule
        Task & Transcription \\
        \midrule
        \textit{HTR} & \texttt{dit dia rebere de Jua Oliveres pages de Llissa demunt viudo ab Maria donsella filla de Juan Pruna pages del far y de Beneta}\\
        \midrule
        \textit{HTR+NER} & \texttt{dit dia rebere de \color{blue}<N-H>\color{black}Jua \color{darkblue}<SN-H>\color{black}Oliveres \color{teal}<O-H>\color{black}pages de \color{purple}<L-H>\color{black}Llissa demunt \color{green}<S-H>\color{black}viudo ab \color{darkred}<N-W>\color{black}Maria \color{red}<S-W>\color{black}donsella filla de \color{yellow}<N-WF>\color{black}Juan \color{orange}<SN-WF>\color{black}Pruna \color{brown}<O-WF>\color{black}pages del \color{pink}<L-WF>\color{black}far y de \color{grey}<N-WM>\color{black}Beneta}\\
        \midrule
        \textit{Key-value HTR+NER} & \texttt{\color{blue}<N-H>\color{black}Jua \color{darkblue}<SN-H>\color{black}Oliveres \color{teal}<O-H>\color{black}pages \color{purple}<L-H>\color{black}Llissa \color{green}<S-H>\color{black}viudo \color{darkred}<N-W>\color{black}Maria \color{red}<S-W>\color{black}donsella \color{yellow}<N-WF>\color{black}Juan \color{orange}<SN-WF>\color{black}Pruna \color{brown}<O-WF>\color{black}pages \color{pink}<L-WF>\color{black}far \color{grey}<N-WM>\color{black}Beneta}\\
        \bottomrule
    \end{tabular}

\end{table}

\section{Experimental results}

In this section, we introduce the evaluation metrics and present the results obtained on each dataset. We also compare our work with state-of-the-art methods and discuss the results.

\subsection{Metrics}

For all three datasets, performances are evaluated by the same standard character recognition and entity recognition metrics, as detailed in the following paragraphs. An additional metric is used to evaluate the experiments on ESPOSALLES.

\subsection{HTR metrics}

The quality of handwriting recognition is evaluated using the character error rate (CER) and word error rate (WER). The full text is evaluated, and named entity tokens are ignored in integrated methods at this step of the evaluation. 

\subsection{NER metrics}

We use the Nerval\footnote{\url{https://gitlab.com/teklia/ner/nerval}} evaluation toolkit to evaluate named entity recognition results. 
In Nerval\cite{miret2021}, the automatic transcription is aligned with the ground truth at character level. Predicted and ground truth words are considered a match if their edit distance is less than 30\%. 
From this alignment, precision, recall and F1-score are computed. 

\subsection{IEHHR metrics}
Finally, for the ESPOSALLES dataset, we also compute the IEHHR metric that was introduced in the ICDAR 2017 Competition on Information Extraction in Historical Handwritten \cite{iehhr2017}. This metric jointly evaluates HTR and NER. Only words associated with named entities are taken into account in this evaluation.
The “basic” score is equal to 100-CER if the \textit{category} tag is correct, 0 otherwise.
The “complete” score is equal to 100-CER if both the \textit{category} and \textit{person} tags are correct, 0 otherwise.

\subsection{Evaluation results}

We present handwritten text recognition results in Table \ref{tab:htr_results} and named entity recognition results in Table \ref{tab:ner_results}. 
For ESPOSALLES, we also provide the results for information extraction in Table \ref{tab:iehhr_esposalles} and obtain state-of-the-art results on the public IEHHR benchmark\footnote{\url{https://rrc.cvc.uab.es/?ch=10&com=evaluation&task=1}}.

\begin{table}[]
    \caption{Evaluation results for handwritten text recognition on IAM, ESPOSALLES, and POPP. Results are given for test sets. NER tokens are not taken into account for this evaluation.}
    \label{tab:htr_results}
    \centering
    \begin{subtable}{0.8\textwidth}
        \caption{IAM (RWTH split)}
        \centering
        \begin{tabular}{llccr}
        \toprule
        Model & Task & CER (\%) & WER (\%) & Input\\ 
        \midrule
        VAN \cite{VAN} & \textit{HTR} & 4.45 & 14.55  & Line \\
        \midrule
        PyLaia & \textit{HTR} & 7.79 & 24.73 & Line \\
        DAN & \textit{HTR} & \textbf{4.30} & \textbf{13.66} &  Page \\
        \midrule
        DAN & \textit{HTR+NER} & 5.12 & 16.17 & Line \\
        DAN & \textit{HTR+NER} & 4.82 & 14.61 &  Page\\
         \bottomrule
         \end{tabular}
        \label{tab:htr_iam}
    \end{subtable}
    \begin{subtable}{0.8\textwidth}
        \centering
        \caption{ESPOSALLES}
        \begin{tabular}{llccr}
        \toprule
        Method & Task & CER (\%) & WER (\%) & Input \\ 
        \midrule
        Seq2seq \cite{Tarride2022-IEHHR-Attention} & \textit{HTR} & 2.82 & 8.33  & Line \\ 
        Seq2seq \cite{Tarride2022-IEHHR-Attention} & \textit{HTR+NER} & 1.81 & 6.10 & Line \\ 
        \midrule
         PyLaia & \textit{HTR} & 0.76 & 2.62 & Line \\
         DAN & \textit{HTR} & 0.46 & \textbf{1.37} & Record \\
         \midrule
         DAN & \textit{HTR+NER}  & 0.48 	& 1.75 & Line \\
         DAN & \textit{HTR+NER} & \textbf{0.39} & 1.51 & Record\\
         DAN & \textit{HTR+NER} & 3.61 & 4.23 & Page\\
         \bottomrule
         \end{tabular}
        \label{tab:htr_esposalles}
        \end{subtable}
    \begin{subtable}{\textwidth}
        \caption{POPP}
        \centering
        \begin{tabular}{llccr}
        \toprule
        Model & Task & CER (\%) & WER (\%) & Input \\ 
        \midrule
        VAN \cite{Constum-POPP} & \textit{HTR} & \textbf{7.08} & 19.05 & Line \\
        \midrule
         PyLaia & \textit{HTR} & 17.19 & 37.43 & Line \\
         DAN & \textit{HTR} & 8.18 & \textbf{18.09}  &  Line \\
         \midrule
         DAN & \textit{HTR+NER} & 7.83 & 24.57 & Line \\
         DAN & \textit{HTR+NER} & 11.74 & 30.78 &  Page\\
         \bottomrule
         \end{tabular}
        \label{tab:popp_ner}
    \end{subtable}
\end{table}

\begin{table}[]
    \caption{Evaluation results for named entity recognition on IAM, ESPOSALLES, and POPP. Results are given for test sets. Evaluation results are computed using Nerval, which computes an alignment between ground truth and predicted entities.}
    \label{tab:ner_results}
    \centering
    \begin{subtable}{\textwidth}
    \centering
    \caption{IAM (RWTH split)}
    \begin{tabular}{lcccr}
    \toprule
        Method & P (\%) & R (\%) & F1 (\%) &  Input Type \\  
        \midrule
        Tülselmann et al.* \cite{Tuselmann2021-NER-seq} & 60.4 & 50.9 & 54.2 & Word/Record\\
        Rowtula et al.* \cite{Rowtula2018-POS}& 33.8 & 30.9 & 32.3 & Word/Record \\
        Todelo et al.* \cite{Toledo2019-context}  & 26.4 & 10.8 & 14.9 & Word/Record\\
        Dessurt \cite{Dessurt} & - & - & 40.4 & Page\\
        \midrule
        Ground-truth + SpaCy & 74.9 & 76.2  & 75.5  & -/Page  \\
        PyLaia + SpaCy & 56.5 & 49.0 & 52.5 & Line/Page \\
        DAN + SpaCy & \textbf{61.8} & \textbf{57.9} & \textbf{59.8} & Page/Page \\
        \midrule
        DAN & 37.1 & 30.8 &	33.7 & Line  \\
        DAN & 37.2 & 27.0 &	31.3 & Page \\
        \midrule
        DAN & 0 & 0 & 0 & Page (key-value) \\
        \bottomrule
        \multicolumn{5}{l}{\small * Different computation method due to pre-existing word alignment.} \\
    \end{tabular}
    \label{tab:nerval_iam}
\end{subtable}
    \begin{subtable}{\textwidth}
    \centering
    \caption{ESPOSALLES}
   \begin{tabular}{llccclcccr}
    \toprule
        Method & \phantom{sep} & \multicolumn{3}{c}{Person} & \phantom{sep} & \multicolumn{3}{c}{Category}  &  Input Type \\  
        & & P (\%) & R (\%) & F1 (\%) & & P (\%) & R (\%) & F1 (\%) & \\  
        \midrule
        Tülselmann et al.* \cite{Tuselmann2021-NER-seq} & & \textbf{99.3} & \textbf{99.2} & \textbf{99.3} & & \textbf{98.5} & \textbf{98.2} & \textbf{98.3} & Word/Record \\
        Rowtula et al.* \cite{Rowtula2018-POS}& & 97.0 & 96.2 & 96.6 & & 97.1 & 97.0 & 97.0 & Word/Record \\
        Todelo et al.* \cite{Toledo2019-context}& & 98.5 & 97.8&  98.1 && 98.5 & 97.8 & 98.1 & Word/Record \\
        \midrule
        Ground-truth + SpaCy  & & 98.6 & 98.4 & 98.5 & & 98.3 & 98.7 & 98.5 & -/Record \\ 
        PyLaia + SpaCy & & 95.9 & 94.0 & 94.9& & 95.6 & 94.3 & 95.0  & Line/Record \\ 
        DAN + SpaCy & &\textbf{97.9} & 97.9 & 97.9 & & \textbf{97.6} & \textbf{98.1} & \textbf{97.8} &  Record/Record \\ 
        \midrule
        DAN & &  96.0 & 96.1 & 96.1 & & 96.9 & 97.0 & 96.9&  Line \\ 
        DAN & &  \textbf{97.9} & 98.2 & \textbf{98.1} & & 97.4 & 97.8 & 97.6 &   Record \\ 
        DAN & & 95.0 & \textbf{98.4} & 96.6 & & 94.2 & 97.6 & 95.9 &  Page \\ 
        \midrule
        DAN & & 97.0 & 97.4 & 97.2 & & 96.7 & 97.1 & 96.9 	 &  Record (key-value) \\ 
        \bottomrule
        \multicolumn{10}{l}{\small * Different computation method due to pre-existing word alignment.} \\
    \end{tabular}
    \label{tab:ner_esposalles}
    \end{subtable}
   
    \footnotetext[1]{Footnote}
    \label{tab:nerval_esposalles}

\begin{subtable}{\textwidth}
    \centering
    \caption{POPP}
    \begin{tabular}{lcccr}
    \toprule
        Method  & P (\%) & R (\%) & F1 (\%) & Input type \\  
        \midrule
        Ground-truth + SpaCy & 95.6 & 97.3 & 96.4 & -/Line  \\
        PyLaia + SpaCy & 75.6 & 77.0 & 76.3 & Line/Line \\
        DAN + SpaCy & 82.8 & 85.3 & 84.0 & Line/Line \\
        \midrule
        DAN & \textbf{85.6} & 86.2 & \textbf{85.9} & Line \\
        DAN & 83.8 & \textbf{86.9} & 85.3 & Page \\
    \bottomrule
    \end{tabular}
    \label{tab:nerval_popp}
\end{subtable}
\end{table}

\begin{table}[]
    \centering
    \caption{IEHHR scores given for the test set of ESPOSALLES dataset.}
    \begin{tabular}{lccr}
    \toprule
    Method & Basic (\%) & Complete (\%) &  Input Type \\ 
    \midrule
    Baseline HMM \cite{iehhr2017} & 80.28 & 63.11 & Line/Line \\ 
    CITlab ARGUS-1 \cite{iehhr2017} & 89.54 & 89.17 & Line/Line \\ 
    CITlab ARGUS-2 \cite{iehhr2017}& 91.63 & 91.19 & Line/Line \\ 
    CITlab ARGUS-3 \cite{iehhr2017}& 91.94 & 91.58 & Line/Line \\
    CVC \cite{Toledo2019-context} & 90.59 & 89.40 & Line/Line \\ 
    Naver Labs \cite{Prasad2018-IEHHR-benchmark} & 95.46 & 95.03 & Line/Line \\ 
    IRISA \cite{Tarride2022-IEHHR-Attention} & 94.7 & 94.0 & Line  \\
    IRISA multi-task \cite{Tarride2022-IEHHR-Attention} &  95.2 & 94.4 & Line  \\
    InstaDeep GNN/Transformer\footnote{Not published yet} & 96.22 & 96.24 & Record \\ 
    InstaDeep Transformer \cite{Rouhou2021-IEHHR-Transformer} & 96.25 & 95.54 & Record \\ 
    TEKLIA Kaldi + Flair \cite{Monroc2022-NER-opensource}&  96.96 & - & Line/Record \\ 
    \midrule
    Ground-truth + SpaCy & 97.51 & 97.57 & -/Record \\
    PyLaia + SpaCy & 96.58 & 96.58 & Line/Record \\ 
    DAN + SpaCy & \textbf{97.13} & \textbf{97.11} & Record/Record \\ 
    \midrule
    DAN  & 96.26 & 94.47 & Line \\ 
    DAN & 97.03 &	96.93 &  Record \\ 
    DAN & 95.45 & 95.04 &  Page \\ 
    \midrule
    DAN (key-value)  & 96.48 &	96.31 & Record (key-value) \\ 
    \bottomrule
    \end{tabular}
    \label{tab:iehhr_esposalles}
\end{table}

\subsubsection{What is the best model for HTR?} Results in Table \ref{tab:htr_results} show that DAN is better than PyLaia for HTR on all three datasets. The DAN model trained only for \textit{HTR} is generally better than the model directly trained for \textit{HTR+NER}. 
The results show that DAN is always better than PyLaia for handwriting recognition: CER and WER are always lower with DAN. 
The WER reaches 1.37\% on ESPOSALLES, 13.66\% on IAM, and 18.09\% on POPP.
Finally, we note that DAN can be more performant on larger text zones. Indeed, DAN performs better on pages on IAM, and on records on ESPOSALLES. On the other hand, on POPP, the best performances are obtained on text lines. This observation can be explained by the fact that POPP documents are tables in which the lines are independent.

\subsubsection{What is the impact of HTR errors on NER?} Results in Table \ref{tab:ner_results} help us understand the impact of handwriting recognition errors on NER performance. The second block of each subtable compares the results using ground transcription or predicted transcriptions (PyLaia or DAN). 
On ESPOSALLES, both HTR systems are very performant with CER below 1\%. As a result, NER performance remains very good. 
However, on IAM and POPP, PyLaia and DAN yield a higher CER. As a consequence, the F1 score drops by 15 points for a 5\% CER on IAM, and by 10 points for a 10\% CER on POPP. 


\subsubsection{What is the best approach for information extraction?}
The best performance on IAM is achieved with a two-stage method, combining DAN (HTR) and SpaCy (NER). These results support the observations of Tüselmann et al. \cite{Tuselmann2021-NER-seq}, and can be explained because there are few entities in the dataset. As a result, DAN struggles to learn semantic information, while SpaCy benefits from pre-trained embeddings for the English language. 
However, on POPP, DAN trained for \textit{HTR+NER} outperforms the two-stage approach combining DAN and SpaCy, although SpaCy does benefit from pre-trained French embeddings. There are two possible explanations for this result. First, POPP documents contain mostly names and surnames, which may not be included in the embeddings. Second, since these are tabular documents, word localization determines the semantic category, as each column corresponds to a specific named entity. Unlike DAN, SpaCy does not have any information regarding the word localization.
Finally, on ESPOSALLES, both approaches yield similar results: SpaCy recognizes the \textit{category} labels better while DAN recognizes the \textit{person} labels better.

\subsubsection{What is the performance of segmentation-free methods?}
It is interesting to note that DAN often performs better on pages (IAM) or records (ESPOSALLES) than on text lines. And yet, the text recognition task is traditionally done on text lines, which requires prior automatic or manual segmentation. But manual segmentation is time-consuming, and automatic segmentation can introduce many errors that affect the performance of handwriting or named entity recognition \cite{Monroc2022-NER-opensource}. 
Therefore, results presented on pages cannot be directly compared to the results on text lines or records, as the task is much harder. 
In order to compare these results fairly, segmentation-based workflows should be evaluated on automatically segmented text lines or records. It is likely that segmentation-free workflows will outperform segmentation-based workflows in an end-to-end evaluation setting.

\subsubsection{Is DAN able to learn from key-value annotations?}
Finally, we evaluate the ability of DAN to learn from key-value annotations. 
On ESPOSALLES, where 50\% of words are linked to an entity, DAN manages to learn from key-value annotations. It learns to recognize relevant words and to ignore the others. Although its performances are slightly lower than when trained with full transcripts, they remain very competitive.
In contrast, DAN fails to learn on IAM, in which only 10\% of words are linked to an entity. The model can be trained for a few epochs before overfitting. As a result, it does not predict anything on the test set. 
Finally, on POPP, all words are linked to an entity, so this experiment is similar to the one with full annotations, as there are no words to ignore during training.

\section{Conclusion}
In this paper, we focus on information extraction in digitized handwritten documents. 
We compare an integrated approach trained for joint HTR and NER with a traditional two-stage approach that performs HTR before NER. We present results at different levels: pages, paragraphs and lines and reach state-of-the-art performance on three datasets.

Our experiments show that integrated approaches trained jointly for HTR and NER can outperform two-stage approaches when word localization has an impact on the NER label (POPP). As opposed, two-stage approaches are better when applied on datasets with few entities (IAM) as they can benefit from pre-trained embeddings. In other cases (ESPOSALLES), two-stage and integrated approaches reach similar performance 97.11\% and 96.93\% respectively, for the complete IEHHR score on records on ESPOSALLES. 
We also demonstrate that applying these models directly on pages leads to very acceptable performances, either better than when applied on lines (ESPOSALLES, IAM), or with a minor performance loss (POPP). The interest of this method is enhanced by the lack of need for prior automatic segmentation, which is known to impact handwriting recognition performances  \cite{Monroc2022-NER-opensource}.
Finally, we show that, under certain conditions, integrated methods are able to learn from key-value annotations, e.g. from a list of relevant words with their corresponding named entities. On ESPOSALLES, the model trained on key-value annotations reaches a complete recognition score of 96.31\%. This observation is encouraging as it would allow training models from incomplete information manually, which considerably reduces the effort needed for manual transcription. 

In future works, we are interested is measuring the impact of segmentation errors when evaluating end-to-end systems for information extraction. 
We also would like to identify the conditions needed to train a model on key-value annotations.
Finally, we want to improve DAN for the task of information extraction. For example, the training loss could also be adapted to differentiate NER tokens from characters. 
Performance could be improved by using pre-trained embeddings like in SpaCy. Since DAN and SpaCy rely on character- and word-embeddings respectively, it would be interesting to find a common representation at sub-word level. 



\bibliographystyle{splncs04}
\bibliography{main}
\section*{Appendix}

\subsection*{Detailed splits for IAM and Esposalles}
We provide the detailed splits used for IAM in Table \ref{tab:iam_stats} and ESPOSALLES in Table \ref{tab:stats_esposalles}. For IAM, we use the RWTH split. For ESPOSALLES, we use the official split, with 25\% of training data used for validation.

\begin{table}[]
    \centering
    \caption{Statistics of the IAM dataset (RWTH split)}
    \label{tab:iam_stats}
    \begin{subtable}{0.42\linewidth}
        \centering
        \caption{Pages, lines, words, and entities by split}
             \begin{tabular}{lrrr}
            \toprule
             & Train & Validation & Test \\
             \midrule
             Pages & 747 & 116 & 336 \\
             Lines & 6,482 & 976 & 2,915 \\
             Words & 55,111 & 8,900 & 25,931 \\
             Entities & 5,868 & 654 & 1,713 \\
             \bottomrule
        \end{tabular}
    \end{subtable}
    \hfill
    \begin{subtable}{0.49\linewidth}
        \centering
        \caption{Entities by split}
            \begin{tabular}{lrrr}
            \toprule
            & Train & Validation & Test \\
            \midrule
            Person & 1,399 & 252 & 603 \\
            GPE & 731 & 38 & 129 \\
            Organization & 825 & 39 & 100 \\
            NORG & 282 & 19 & 79 \\
            Date & 1,000 & 57 & 178 \\
            Cardinal & 409 & 75 & 130 \\
            Work of Art & 294 & 41 & 110 \\
            Time & 167 & 24 & 114 \\
            FAC & 126 & 37 & 71 \\
            Quantité & 107 & 17 & 66 \\
            Location & 124 & 16 & 41 \\
            Ordinal & 104 & 19 & 38 \\
            Product & 78 & 6 & 24 \\
            Percent & 91 & 6 & 4 \\
            Event & 61 & 2 & 15 \\
            Law & 43 & 6 & 0 \\
            Language & 15 & 0 & 5 \\
            Money & 12 & 0 & 6 \\
           \bottomrule
        \end{tabular}
    \end{subtable}
\end{table}

\begin{table}[]
    \centering
    \caption{Statistics of the ESPOSALLES dataset}
    \label{tab:stats_esposalles}
    \begin{subtable}{0.42\linewidth}
        \centering
        \caption{Pages, records, lines, words, and entities by split}
             \begin{tabular}{lrrr}
            \toprule
             & Train & Validation & Test \\
             \midrule
             Pages & 75 & 25 & 25 \\
             Records & 731 & 267 & 253 \\
             Lines & 2,328 & 742 & 757 \\
             Words & 23,893 & 7,608 & 8,026 \\
             Entities & 12,388 & 3,937 & 4,238 \\
             \bottomrule
        \end{tabular}
    \end{subtable}
    \hfill
    \begin{subtable}{0.49\linewidth}
        \centering
        \caption{Entities by split}
            \begin{tabular}{lrrr}
            \toprule
            & Train & Validation & Test \\
            \midrule
            Name & 3,774 & 1,223 & 1,312 \\
            Surname & 2,033 & 634 & 694 \\
            Location & 3,440 & 1,069 & 1,087 \\
            Occupation & 2,273 & 737 & 797 \\
            State & 868 & 274 & 319 \\
            \midrule
            Wife & 2,093 & 678 & 768 \\
            Wife's father & 2,745 & 847 & 908 \\
            Wife's mother & 566 & 188 & 189 \\
            Husband & 4,334 & 1,493 & 1,563 \\
            Husband's father & 1,838 & 476 & 518 \\
            Husband's mother & 462 & 1401 & 156 \\
            Other person & 350 & 115 & 136 \\
           \bottomrule
        \end{tabular}
    \end{subtable}
\end{table}

\subsection*{Impact of curriculum learning}

We evaluate the impact of curriculum learning for the task of \textit{HTR+NER} in Table \ref{tab:curriculum_random_htr} and \ref{tab:curriculum_random_ner}. The DAN model trained with curriculum learning is pre-trained on the \textit{HTR} task, then fine-tuned on the \textit{HTR+NER} task.
The results show that curriculum learning does not always have a positive impact on final performances.

For POPP, we also trained the model for key-value \textit{HTR+NER} in a random order, e.g. with named entities in a random order. Results show that DAN is also able to learn with a random reading order, although the error rates are a bit higher than when the model is trained with the correct reading order.

\begin{table}[]
    \caption{Impact of curriculum learning on handwritten text recognition on IAM, ESPOSALLES, and POPP. Results are given for test sets. NER tokens are not taken into account for this evaluation.}
    \label{tab:curriculum_random_htr}
    \centering
    \begin{subtable}{0.8\textwidth}
        \caption{IAM (RWTH split)}
        \centering
        \begin{tabular}{llccr}
        \toprule
        Model & Task & CER (\%) & WER (\%) & Input\\ 
        \midrule
        DAN & \textit{HTR} &  4.86 & 15.78 &  Line \\
        DAN & \textit{HTR} & \textbf{4.30} & \textbf{13.66} &  Page \\
        \midrule
        DAN & \textit{HTR+NER} & 5.12 & \textbf{16.17} & Line \\
        DAN curriculum & \textit{HTR+NER} & \textbf{5.01} & 16.32 & Line \\
        \midrule
        DAN & \textit{HTR+NER} & 4.82 & 14.61 &  Page\\
        DAN curriculum & \textit{HTR+NER}  & \textbf{4.30} & \textbf{13.65} & Page \\
         \bottomrule
         \end{tabular}
    \end{subtable}
    \begin{subtable}{0.8\textwidth}
        \centering
        \caption{ESPOSALLES}
        \begin{tabular}{llccr}
        \toprule
        Method & Task & CER (\%) & WER (\%) & Input \\ 
        \midrule
         DAN & \textit{HTR} & 0.54 & 2.13  & Line \\
         DAN & \textit{HTR} & 0.46 & \textbf{1.37} & Record \\
         DAN & \textit{HTR} & 2.77 & 3.58 & Page \\
         \midrule
         DAN & \textit{HTR+NER}  & \textbf{0.48} 	& \textbf{1.75} & Line \\
         DAN curriculum & \textit{HTR+NER} & 0.64 & 2.02 & Line \\
         \midrule
         DAN & \textit{HTR+NER} & \textbf{0.39} & \textbf{1.51} & Record\\
         DAN curriculum & \textit{HTR+NER} & 0.89 & 1.97& Record \\
         \midrule
         DAN & \textit{HTR+NER} & 3.61 & 4.23 & Page\\
         DAN curriculum & \textit{HTR+NER} & \textbf{2.23} & \textbf{3.15}  & Page \\
         \bottomrule
         \end{tabular}
        \end{subtable}
    \begin{subtable}{\textwidth}
        \caption{POPP}
        \centering
        \begin{tabular}{llccr}
        \toprule
        Model & Task & CER (\%) & WER (\%) & Input \\ 
        \midrule
         DAN & \textit{HTR} & 8.18 & \textbf{18.09}  &  Line \\
         \midrule
         DAN & \textit{HTR+NER} & \textit{7.83} & \textit{24.57} & Line \\
         DAN curriculum & \textit{HTR+NER} & 8.06 & 24.85 & Line \\
         DAN curriculum + random order & \textit{HTR+NER} & 9.53  & 27.01 & Line \\
         \bottomrule
         \end{tabular}
    \end{subtable}
\end{table}

\begin{table}[]
    \caption{Impact of curriculum learning for named entity recognition on IAM, ESPOSALLES, and POPP. Results are given for test sets. Evaluation results are computed using Nerval, which computes an alignment between ground truth and predicted entities.}
    \label{tab:curriculum_random_ner}
    \centering
    \begin{subtable}{\textwidth}
    \centering
    \caption{IAM (RWTH split)}
    \begin{tabular}{lcccr}
    \toprule
        Method & P (\%) & R (\%) & F1 (\%) &  Input Type \\  
        \midrule
        DAN & 37.1 & 30.8 &	33.7 & Line  \\
        DAN curriculum & 33.0 & 23.3 & 27.3  & Line \\
        \midrule
        DAN & 37.2 & 27.0 &	31.3 & Page \\
        DAN curriculum & 38.2 & 29.1 & 33.1 & Page \\
        \bottomrule
    \end{tabular}
\end{subtable}
    \begin{subtable}{\textwidth}
    \centering
    \caption{ESPOSALLES}
   \begin{tabular}{llccclcccr}
    \toprule
        Method & \phantom{sep} & \multicolumn{3}{c}{Person} & \phantom{sep} & \multicolumn{3}{c}{Category}  &  Input Type \\  
        & & P (\%) & R (\%) & F1 (\%) & & P (\%) & R (\%) & F1 (\%) & \\  
        \midrule
        DAN & &  96.0 & 96.1 & 96.1 & & 96.9 & 97.0 & 96.9&  Line \\ 
        DAN curriculum & & 95.6 & 94.0 & 94.8 & & 96.3 & 95.5 & 95.9 &  Line \\ 
        \midrule
        DAN & &  \textbf{97.9} & 98.2 & \textbf{98.1} & & 97.4 & 97.8 & 97.6 &   Record \\ 
        DAN curriculum & &  97.3 & 97.5 & 97.4 & & 96.5 & 97.3 & 96.9 	&   Record \\ 
        \midrule
        DAN & & 95.0 & \textbf{98.4} & 96.6 & & 94.2 & 97.6 & 95.9 &  Page \\ 
        DAN curriculum  & & 96.4 & 97.3 & 96.9 & & 95.4 & 97.2 & 96.3 &  Page \\ 
        \midrule
        DAN & & 97.0 & 97.4 & 97.2 & & 96.7 & 97.1 & 96.9 	 &  Record (key-value) \\ 
        DAN curriculum & & 96.7 & 96.3  & 96.5 & & 96.0 & 96.1 & 96.0 &  Record (key-value) \\ 
        \bottomrule
    \end{tabular}
    \end{subtable}
   
\begin{subtable}{\textwidth}
    \centering
    \caption{POPP}
    \begin{tabular}{lcccr}
    \toprule
        Method  & P (\%) & R (\%) & F1 (\%) & Input type \\  
        \midrule
        DAN & \textbf{85.6} & \textbf{86.2} & \textbf{85.9} & Line \\
        DAN curriculum  &  85.4 & \textbf{86.2} & 85.8 & Line \\
        DAN curriculum + random order & 84.6 & 84.8 & 84.7 & Line\\
    \bottomrule
    \end{tabular}
\end{subtable}
\end{table}
\end{document}